# Real-World Federated Learning in Radiology: Hurdles to overcome and Benefits to gain

Markus R. Bujotzek*[,1,2], Ünal Akünal[1], Stefan Denner[1,2], Peter Neher[1,3,4], Maximilian Zenk[1,2], Eric Frodl[5], Astha Jaiswal[6], Moon Kim[7], Nicolai R. Krekiehn[8], Manuel Nickel[9], Richard Ruppel[10], Marcus Both[11], Felix Döllinger[10], Marcel Opitz[13], Thorsten Persigehl[6], Jens Kleesiek[7], Tobias Penzkofer[10], Klaus Maier-Hein[1,2,3,4,14], Rickmer Braren[†,15], Andreas Bucher[†,5]

[1] Division of Medical Image Computing, German Cancer Research Center (DKFZ), 69120 Heidelberg, Germany

[2] Medical Faculty Heidelberg, University of Heidelberg, 69120 Heidelberg, Germany

[3] Pattern Analysis and Learning Group, Department of Radiation Oncology, 69120 Heidelberg University Hospital, Heidelberg, Germany

[4] German Cancer Consortium (DKTK), partner site Heidelberg, 69120 Heidelberg, Germany

[5] Institute for Diagnostic and Interventional Radiology, University Hospital Frankfurt; Goethe University Frankfurt; 60590 Frankfurt, Germany

[6] Institute for Diagnostic and Interventional Radiology, Faculty of Medicine and University Hospital Cologne, University of Cologne, 50937 Cologne, Germany

[7] Institute for AI in Medicine (IKIM), University Hospital Essen (AöR), 45131 Essen, Germany

[8] Intelligent Imaging Lab@Section Biomedical Imaging, Dept. of Radiology and Neuroradiology, University Medical Center Schleswig-Holstein (UKSH), Campus Kiel, Germany

[9] Institute for AI in Medicine, Technical University of Munich, 81675 Munich, Germany

[10] Department of Radiology, Charité - Universitätsmedizin Berlin, 10117 Berlin, Germany

[11] Department of Radiology and Neuroradiology, University Medical Centers Schleswig-Holstein, Campus Kiel, Kiel, Germany

[12] Department of Radiology, Charité - Universitätsmedizin Berlin, 10117 Berlin, Germany

[13] Institute for Diagnostic and Interventional Radiology and Neuroradiology, University Hospital Essen (AÖR), 45131 Essen, Germany

[14] National Center for Tumor Diseases (NCT), NCT Heidelberg, a partnership between DKFZ and the university medical center Heidelberg, Heidelberg, Germany

[15] Institute for diagnostic and interventional radiology, Klinikum rechts der Isar, Technical University of Munich, 81675 Munich, Germany

* markus.bujotzek@dkfz-heidelberg.de

[†] contributed equally

## Abstract

**Objective:** Federated Learning (FL) enables collaborative model training while keeping data locally. Currently, most FL studies in radiology are conducted in simulated environments due to numerous hurdles impeding its translation into practice. The few existing real-world FL initiatives rarely communicate specific measures taken to overcome these hurdles, leaving behind a significant knowledge gap. Minding efforts to implement real-world FL, there is a notable lack of comprehensive assessment comparing FL to less complex alternatives.

**Materials & Methods:** We extensively reviewed FL literature, categorizing insights along with our findings according to their nature and phase while establishing a FL initiative, summarized to a comprehensive guide. We developed our own FL infrastructure within the German Radiological Cooperative Network (RACOON) and demonstrated its functionality by training FL models on lung pathology segmentation tasks across six university hospitals. We extensively evaluated FL against less complex alternatives in three distinct evaluation scenarios.
**Results:** The proposed guide outlines essential steps, identified hurdles, and proposed solutions for establishing successful FL initiatives conducting real-world experiments. Our experimental results show that FL outperforms less complex alternatives in all evaluation scenarios, justifying the effort required to translate FL into real-world applications.
**Discussion & Conclusion:** Our proposed guide aims to aid future FL researchers in circumventing pitfalls and accelerating translation of FL into radiological applications. Our results underscore the value of efforts needed to translate FL into real-world applications by demonstrating advantageous performance over alternatives, and emphasize the importance of strategic organization, robust management of distributed data and infrastructure in real-world settings.

**Key words:** radiology, artificial intelligence, federated learning, healthcare infrastructure, distributed systems

## OBJECTIVES

Deep learning (DL) has revolutionized radiological image analysis. It is rapidly driving innovation in radiological research and increasingly transforming clinical routine. Training powerful DL models requires access to vast and diverse datasets. A practical approach is to train on centralized data from multiple centers in a pooled data lake. However, such data aggregation is often complicated by various overlapping or even conflicting regulatory requirements, including privacy regulations such as GDPR and HIPAA, or state-specific healthcare laws and federal privacy laws[1]. Federated Learning (FL)[2] resolves these issues by allowing data to remain at originating medical centers. FL derived models are collaboratively trained through periodic exchanges of model weights, achieving performances comparable to centrally trained models[3].
Most FL research is currently conducted in simulated environments[4] lacking broad translation into the real-world applications due to practical implementation challenges. The few studies examining real-world FL in general medicine[5–8] and specifically in radiology[6, 9–14], provide limited insights into real-world challenges of implementing FL in practice, leaving a significant knowledge gap. Additionally, theoretical FL studies[4, 9, 15–30], which discuss hypothetical applications of FL in medicine and potential challenges, often do not account for real-world complexities, reducing their practical relevance to clinical settings.
Within the framework of the German Radiological Cooperative Network (RACOON)[1][31], we have built and established the first nation-wide FL initiative of its kind, that includes all 38 university hospitals of the country. We evaluated its functionality by conducting real-world FL experiments to collaboratively train DL segmentation models on radiological image data across six university hospitals. This setup provides a ready-to-use infrastructure for future researchers to conduct clinical research using FL without having to build their own systems. During developing this FL initiative and conducting experiments with real-world datasets in a real-world setting, we encountered and overcame numerous difficulties and practical hurdles.

[1] https://racoon.network/

Our contributions are threefold. First, we demonstrate the functionality of our FL infrastructure through real-world training of DL segmentation models for lung pathology detection across data distributed over six sites. Second, we propose a detailed guide to build real-world FL initiatives, that merges insights from our first-hand experiences along with relevant literature. This guide outlines essential steps, highlights encountered issues and provides solutions for each phase of building and deploying FL in the real world. Designed to support upcoming real-world FL projects, our guide aims to ensure and accelerate the successful development of future FL initiatives without retracing past missteps. Third, acknowledging the efforts detailed in our proposed guide necessary for real-world FL, we compare FL to less complex alternatives like local model training and ensembling approaches[32–34]. We benchmark these approaches extensively across various evaluation scenarios: personalization, i.e. benefits to gain from joining collaborative training; and generalization with and without local training, i.e. benefits to gain from leveraging collaborative trained models of other sites with and without incorporation of own local training efforts.

# MATERIALS AND METHODS

## Categorization of insights in building real-world FL initiatives

The term “challenges” is frequently used in FL literature, yet the issues encountered in actual real-world implementations of FL initiatives are diverse. We define “real-world” studies as those utilizing clinical trial data rather than curated public datasets and employing distributed computing infrastructure integrated into clinical IT ecosystems. Additionally, we characterize a FL initiative as the combination of community engagement, organizational structures, legal agreements, and infrastructure necessary to conduct and scale FL experiments beyond single execution.

We classify challenges of real-world FL implementation into two categories: practical hurdles and inherent difficulties. *Practical Hurdles* encompass organizational, legal, or technical issues that are solved through agreements or technical solutions. We share our solutions based on practical experiences with these hurdles. In contrast, inherent *difficulties* refer to limitations in real-world FL of organizational, technical, or research-related nature that cannot be avoided but must be acknowledged when successfully developing FL initiatives and conducting real-world experiments.

We further categorize these issues based on the scope within which they impact the establishment of a FL initiative. These categories include organization, legal requirements, infrastructure setup, experiment preparation, and experiments and evaluation.

## Real-world FL study: Experimental setup

### FL infrastructure

Our real-world FL efforts are part of the German RACOON initiative, which aims to use artificial intelligence to advance research in radiological diagnosis and therapy across all 38 German university hospitals. As part of this initiative, each hospital has been equipped with a server hosting key software components: Mint Lesion[2] for structured radiological reporting, SATORI[3] and ImFusion Labels[4] for imaging data annotation, and a Kaapana[5]-based platform for medical image processing. In this setup,

[2] https://mint-medical.com/de/mint-lesion

[3] https://www.mevis.fraunhofer.de/de/research-and-technologies/werkzeuge-fuer-ki-kollaborationen.html

[4] https://www.imfusion.com/products/imfusion-labels

[5] https://www.kaapana.ai/

Kaapana facilitates curating radiological data[35] for subsequent local and federated training[34] of DL models, supporting various studies within RACOON.
In our real-world FL experiments, we orchestrated a centralized FL collaboration of six university hospitals by connecting their Kaapana platforms to a central server. The six university hospitals were: Charité Berlin (CHA), Technical University of Munich (TUM), the University Medicine Essen (UME), and the university hospitals in Frankfurt am Main (UKF), Cologne (UKK) and Kiel (UKKI), see figure 1.

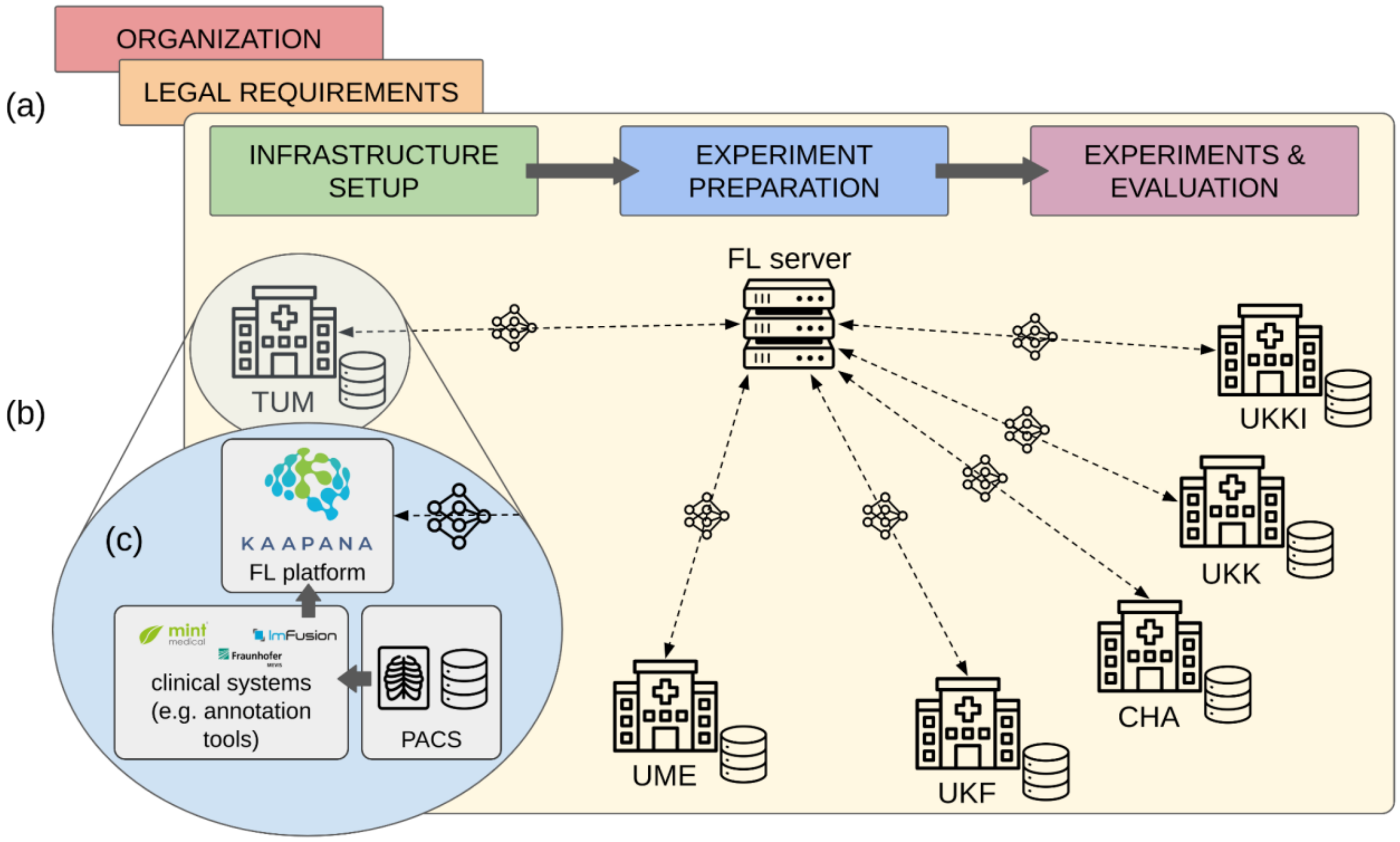

Figure 1: Phases of the proposed guide for building and deploying real-world FL in radiology (a) and the infrastructure of the RACOON FL initiative (b). FL infrastructure with a central FL server and six participating sites (TUM, UME, UKF, CHA, UKK, UKKI) maintaining data locally and periodically exchanging model weights during FL training. Detailed view of the site infrastructure (c): radiological images are queried from the PACS, processed by third party clinical systems (e.g. annotation tools) and used for FL training using the FL platform.

## Dataset

In our FL experiments, we trained a DL model on a distributed dataset of pathological lung CT scans. For this dataset voxel-level segmentation annotations of three pathologies were created by independent radiological readers supervised by experienced board-certified radiologists at each site. The three types of pathological image patterns segmented were: (non-)malignant consolidation (Cons), ground-glass opacity (GGO), and pleural effusion (PE), which are significant predictors of disease progression in various lung diseases[36]. To avoid bias towards specific diseases, the dataset was curated to maintain a balanced number of samples from 20 different lung diseases across all sites. Data provision varied between sites: TUM, UME, and UKF provided manually generated voxel-level annotations; sites CHA, UKK, and UKKI provided automatically pre-processed, manually corrected annotations. For model training, data at each site was split into training and test sets with an 80% to 20% ratio, with special care taken to maintain this ratio for less common PE cases. Following this data-splitting and site-specific inclusion and exclusion criteria, resulted in curated datasets per site as shown in figure2. Detailed descriptive statistics of the distributed data are shown in figure 3.

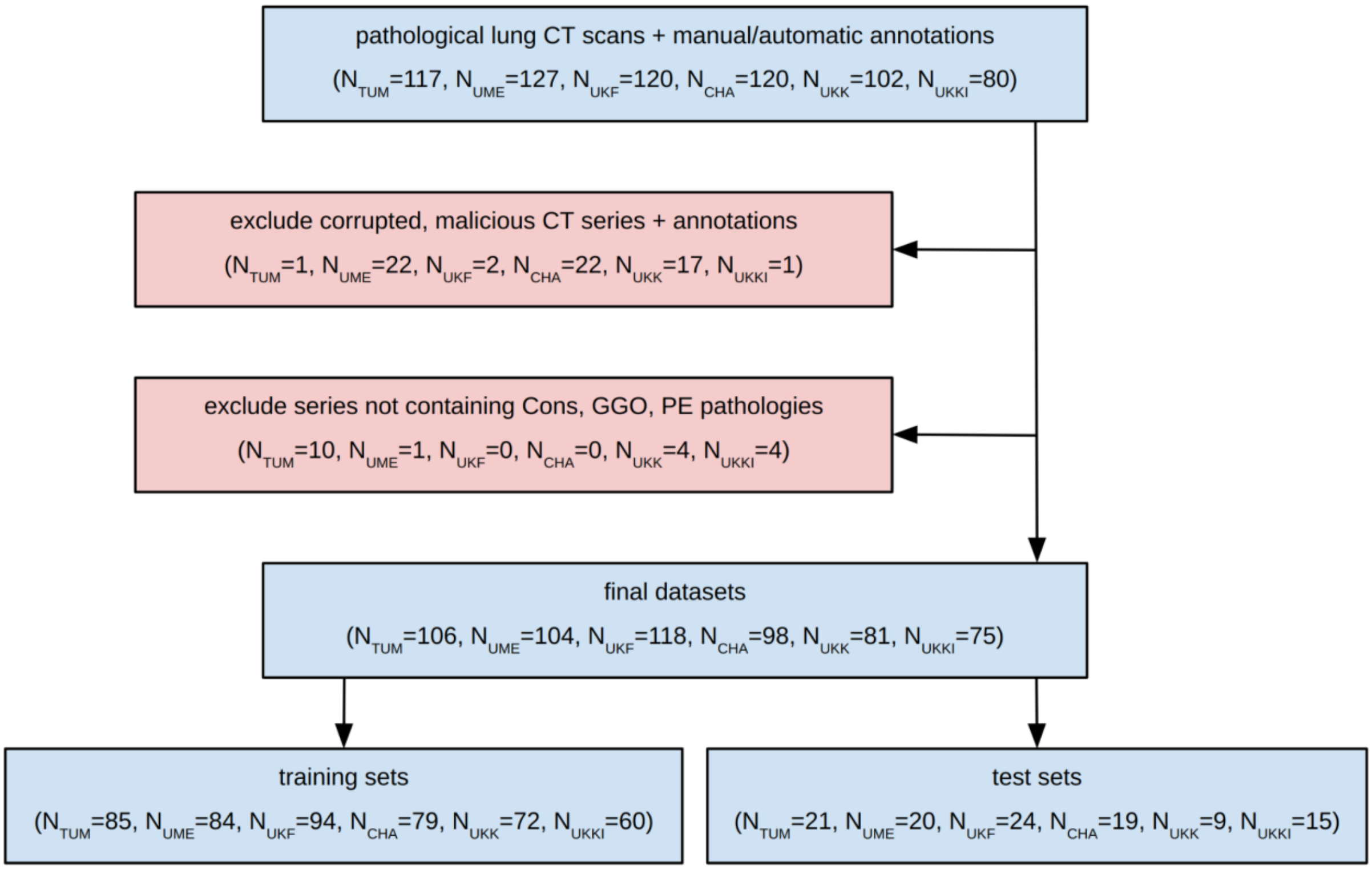

Figure 2: Cohort definition after data curation, filtering and splitting of the distributed data across the six participating sites resulting in the final training and test sets.

## Training details

The DL model utilized is a U-Net model from the self-configuring medical image segmentation framework nnU-Net[37]. The self-configuration process of the nnU-Net model is based on a dataset fingerprint, generated from training data, optimizing the model configuration by rule-based, fixed, and empirical parameter selection.

The model's self-configuration process is straightforward for local training with non-distributed data. However, in a federated setup, this procedure requires multiple steps to synchronize across all participating sites, following the implementation in[34]. Each site generates a dataset fingerprint from its local training data, which is sent to the central server. The server averages these fingerprints and redistributes them to all sites, ensuring each site configures and initializes the model based on this averaged information identically.

In our experiments, we opted for the low-resolution configuration of the nnU-Net model to optimize training efficiency. We stuck to the nnU-Net self-configured parameters without further modifications. Consequently, we trained the model for fixed 1000 epochs (ideal duration for local training[37]), which eliminated the need for a validation set to maximize training sample size. Each local model processed a fixed number of 250 batches per epoch; further nnU-Net training details provided in table A.1. For federated aggregation, we utilized non-weighted averaging updating local model weights $w$ after each local epoch and federated communication round $t$, according to equation 1.

$$w^{(t+1)} = w^{(t)} + \sum_{i=1}^{N_{sites}} \frac{\Delta_i^{(t)}}{N_{sites}} \tag{1}$$

### Evaluation metrics and ranking

To evaluate segmentation performances of trained models, we selected the following metrics according to investigated pathologies[38–40]. We chose the intersection-based Dice Similarity Coefficient (DSC), as it is the default segmentation metric and according to[39, 40] suitable for the three target pathologies. We assessed the segmentation performance using the distance-based metrics Normalized Surface Dice (NSD), especially suitable for Cons and GGO[39] with a threshold of 1 mm, and the Hausdorff Surface Distance (HSD), relevant for PE[40]. We considered the medically relevant difference of predicted and annotated volumes using the Normalized Average Volume Error (NAVE)[39].

Based on the utilized metric implementation[41], we disregard False Positive predictions. For False Negative predictions, we set DSC and NSD to 0.0, HSD to 260.0 mm, the height of a lung[42], and NAVE to 20.0, twice the average of True Positive predictions from local models $L_i$.

As we consider all metrics as equally relevant, we determine the best performing method by combining the metric results in a ranking. We compute for each site and metric the mean metric $\overline{m}$ over all test samples $N_{test}$ and classes $N_{classes}$, resulting in $N_{metrics}$ scores per site. All compared methods are ranked $rank(\cdot)$ per metric score. All $N_{sites} \times N_{metrics}$ rankings are averaged to obtain the overall ranking $r$, see equation 2.

$$r = \frac{1}{N_{sites} \cdot N_{metrics}} \sum_{i=1}^{N_{sites}} \sum_{m=1}^{N_{metrics}} rank(\overline{m}) \tag{2}$$

### Study design

In our experimental studies, we demonstrate the execution of real-world FL experiments within our built FL infrastructure. First, we investigate data characteristics via descriptive statistics assessing the data heterogeneity.

Acknowledging practical hurdles and difficulties of real-world FL, we recognize that radiologists may lean towards less complex alternatives. Following previous studies[32–34], we compare the model performances trained via federated learning ($FL$) versus locally trained models at a single site $i$ ($L_i$) and an ensemble of these local models ($E$). Additionally, we assess specialized versions of the $FL$ and $E$ models, $Spec(FL)$ and $Spec(E)$, by ensembling them with the local model ($L_i$) specific to site $i$ being evaluated.

We evaluate the compared models across three distinct evaluation scenarios summarized in table 1.

The *personalization* scenario evaluates how a participating site can obtain improved models from joining collaborative efforts. Given the heterogeneous annotation procedures, we first investigate personalization capabilities in three-sites experiments with homogeneous annotation procedures each, before extending to six-sites experiments. Thereby, we assess segmentation performance by comparing locally trained models ($L_i$), an ensemble of locally trained models ($E$), a federated trained model ($FL$), and their specialized versions $Spec(E)$ and $Spec(FL)$. Hereinafter, we explore generalization capabilities of models trained across the three sites with manual annotations.

The second scenario explores *generalization without local training*, focusing on sites that cannot train their own models, therefore rely solely on models trained at other sites or through collaborative efforts of those. We benchmark the model performances of all other local models $L_{j \neq i}$, excluding local model $L_i$ while testing on site $i$, against the ensemble of those local models $E_{leave-i-out}$ and a federated trained model excluding site $i$, $FL_{leave-i-out}$.

The third scenario addresses *generalization with local training*, targeting sites that have local training capabilities but are hesitant to join real-world FL efforts due to its complexities. We compare the performances of local models $L_i$, the ensemble of those, $E$, a federated trained model excluding site $i$, $FL_{leave-i-out}$, and the specializations $Spec(E)$ and $Spec(FL_{leave-i-out})$.

Table 1: Comprehensive overview of compared models in the three distinct evaluation scenarios: personalization, generalization with local training, generalization without local training.

| models \ scenario | personalization | generalization without local training | generalization with local training |
|---|---|---|---|
| $L_i$ | ✓ | ✗ | ✓ |
| $L_{j \neq i}$ | ✗ | ✓ | ✗ |
| $E$ | ✓ | ✗ | ✓ |
| $E_{leave-i-out}$ | ✗ | ✓ | ✗ |
| $FL$ | ✓ | ✗ | ✗ |
| $FL_{leave-i-out}$ | ✗ | ✓ | ✓ |
| $Spec(E)$ | ✓ | ✗ | ✓ |
| $Spec(FL)$ | ✓ | ✗ | ✗ |
| $Spec(FL_{leave-i-out})$ | ✗ | ✗ | ✓ |

# RESULTS

## Insights in building real-world FL initiatives

We share our experiences from developing a real-world FL initiative, combined with insights from applied[5–14] and theoretical literature[4, 9, 15–30], to provide the most complete picture. These insights into building a FL initiative and conducting real-world experiments are compiled to a comprehensive guide, navigating through various phases, steps, and issues of translating FL into the real world, see table 2.

Existing works point to practical hurdles in organizing a FL initiative like convincing site's IT departments and governance stakeholders to participate through incentives[5, 14], and the importance of achieving terminology harmonization across the initiative[14]. They also discuss organizational issues, including methods to assess a site's data quantity, quality, heterogeneity, and infrastructure contributions[25], the need for available on-site personnel (IT staff, expert annotators)[22], and strategies for incorporating human oversight, particularly across time zones[21, 22]. There is an emphasis on establishing clear governance, traceability, and accountability for human expertise and site policies[22], and setting technical requirements for the FL platform regarding data access history, training configurations, and error handling[15, 30]. We identified further hurdles including the need to create detailed specifications for radiological imaging data and annotations, establishing effective communication channels like regular meetings and chat rooms among FL researchers, radiologists, and IT staff, and aligning the FL platform's development cycle with necessary features and fixes. Selecting the utilized state-of-the-art DL algorithm and FL aggregation strategy were difficulties to clarify.

Legal steps involve negotiating bidirectional contracts between sites and the FL initiative itself[12], potentially across diverse legal jurisdictions[13], and regulations regarding software support and audits[14]. Beyond that, we faced the question whether model weights could be considered as non-patient-related data.

Practical steps in infrastructure setup include essential on-site infrastructure requirements[4, 16, 17, 21–27, 29, 30], acquiring and connecting hardware, provisioning and accessing on-site virtual machines (VMs)[14], managing limited disk space and resources[4, 8, 14], configuring firewall permissions, and establishing communication with third-party systems, e.g. PACS and annotation tools[4, 14, 24]. Technical difficulties include minimizing the strain on site resources[14] and managing limited communication capacities[8]. Beyond existing literature, we faced significant practical hurdles and difficulties such as installing FL platforms in the highly restricted clinical IT environments and debug sessions between FL platform engineers and local IT to resolve network access and communication issues between the FL platform and third-party systems.

During the experiment preparation phase, practical hurdles primarily involve data-related issues including low quality but high heterogeneity[13, 15, 19, 20, 22, 23, 25, 30], the necessity to standardize custom data[14], addressing missing data harmonization, and inconsistencies[13], and the need for manual inspections due to insufficient data specifications[10]. Additionally, the necessity for on-site debugging possibilities for FL researchers[25] and the trade-off between allowing researchers to find relevant data on-site while maintaining data privacy[15, 22], are pointed out. Similarly, hurdles we faced were handling of data that violated standards or specifications, ensuring data was ready for experiments through validators, and mitigating avoidable data heterogeneity from varying annotation procedures due to insufficient data specifications. Since FL researchers could not directly inspect data, specific annotation inconsistencies were only detected through notably poorer model performance on specific sites.

In the phase of experiments and evaluation, hurdles included managing infrastructure failures, such as site dropouts[5] due to straggling[21], crashing[24, 27, 29], and offline sites[23]. These issues were exacerbated by insufficient error logging and limited technical documentation, necessitating on-site debugging[10]. Technical difficulties involve variation in time per federated communication round due to infrastructure heterogeneity across the initiative[9], the necessity to minimize strain on site's resources[14], managing limited communication capacities[8] and dealing with data heterogeneity that delayed the convergence of federated models[21]. In addition, we encountered hurdles such as failing experiments due to site connection losses, requiring experiment restarts; and technical difficulties inheriting from heterogeneous infrastructure, causing straggling sites and resulting in idle machines at others. Lastly, it is crucial to ensure that the final FL model is readily available on-site for evaluation.

Table 2: Detailed guide to build and conduct real-world FL outlining the phases, steps, and associated issues. Literature insights are cited; our insights are marked with ★.

| **Steps & Issues (D: difficulty, H: hurdle, S: solution)** |
|---|
| **Phase 1: Organization** |
| Step 1: Identify and convince sites with medically and technically motivated governance stakeholders (★,[5, 14])<br>- Hurdle 1: Encourage sites to contribute communication/computation overhead, high-quality data (★,[20, 23])<br>*→Solution 1.1: Assess each site's contribution in terms of data quantity, -quality, -heterogeneity, infrastructure (*[25]*)*<br>*→Solution 1.2: Provide sites incentives, e.g. enhanced performance of FL model, scientific credentials, visibility (*★*)* |
| Step 2: Conclude non-technical agreements covering health protocols, intellectual property ([15]), scientific acknowledgement (★) |
| Step 3: Identify responsible IT personnel and radiologists at participating sites (★)<br>- *H3*: Ensure each site has motivated, capable, and available IT staff and radiologists (★,[22])<br>*→S3: Provide incentives for actual involved persons (IT staff, radiologists), e.g. scientific credentials, visibility (*★,[5, 14]*)* |
| Step 4: Coordinate harmonized acquisition (prospective data) or curation (retrospective data) of radiological images and annotations (★)<br>- H4: Harmonize terminologies within FL initiative (★,[14])<br>*→S4: Develop a detailed data specification and a glossary (*★*)* |
| Step 5: Facilitate communication within FL initiative (★)<br>- H5.1: Ensure regular exchange among directly involved participants (★)<br>*→S5.1: Establish dedicated communication channels such as routine meetings, email, and chat systems (*★*)*<br>- D5.2: Incorporate human input across varying time zones and geographic locations ([21, 22]) |
| Step 6: Select a FL platform (★)<br>- H6: Reach consensus on a unified FL platform used by all sites (★)<br>*→S6: Define requirements, decide using decision-making tools, e.g. pairwise comparison matrix (*★,[15, 22, 30]*)* |
| Step 7: Identify DL algorithm and FL methodology (★)<br>- D7: Select state-of-the-art DL and FL methods (★) |
| Step 8: Identify task to evaluate, e.g. communication efficiency, model performance, security robustness ([16, 17, 26]) |
| Step 9: Determine best-performing method according to chosen evaluation task (★)<br>- H9: Identify suitable evaluation metrics (★)<br>*→S9.1: Choose evaluation metrics from literature solving related problems (*★*)*<br>*→S9.2: follow advising works, e.g.*[38] *for evaluation model performances on medical image data (*★*)* |
| **Phase 2: Legal requirements** |
| Step 1: Conclude contracts between FL initiative and participating sites, e.g. for data processing, ethics (★)<br>- H1: Design legal framework across diverse legal jurisdictions (★,[13])<br>*→S1: Draft individual legal contracts between the FL initiative and each site (*★,[12]*)* |
| Step 2: Develop legal framework for sharing of DL models within the FL initiative (★)<br>- H2: Determine if DL models constitute patient-related data (★)<br>*→S2: Classify shared DL models as non-patient data (*★*)* |
| Step 3: Develop legal framework between FL platform provider and participating sites (★)<br>- H3: Define the scope of support that FL platform engineers can provide on-site (★)<br>*→S3: Conclude contracts allowing on-site support of FL platform engineers (*★*)*<br>- D3: Ensure compliance with each site's software audit requirements ([14]) |
| **Phase3: Infrastructure set-up** |
| Step 1: Acquire, secure, and connect hardware resources on-site (★,[4, 14, 16, 17, 21–27, 29, 30])<br>- D1: Obtain permissions to reserve disk space on limited resources ([8, 11, 14]) |
| Step 2: Provision and access VMs for deploying the FL platform on-site (★,[14]) |
| Step 3: Install the FL platform within the clinical IT ecosystem (★)<br>- H3.1: Sites with highly restricted network access (★)<br>*→S3.1: Offer an offline installation option for the FL platform (*★*)*<br>- D3.2: Prevent site-specific customizations of FL platform affecting e.g. functionality of federated communication (★) |
| Step 4: Configure network settings to allow site VMs to access the container registry and connect to the FL server (★,[14]) |
| Step 5: Locate data in relevant source systems and configure secure read-access ([14]) |
| Step 6: Identify and configure communication endpoints between FL platform and third-party systems (★,[14])<br>- H6: communication issues between FL platform and third-party systems (★) |
| *→S6: Debug communication issues with respective engineers and local IT (★,[4, 24])* |
| Step 7: Manage urgent features requests or bug fixes in FL platform (★)<br>- D7.1: Required features/hotfixes do not necessarily align with software development cycles (★)<br>- H7.2: Coordinate roll-out of FL platform hot-fix releases (★)<br>*→S7.2: Conduct joint sessions between FL platform engineers and local IT (★)*<br>- D7.3: Maintain version compatibility of the FL platform across the initiative (★) |
| **Phase 4: Experiment preparation** |
| Step 1: Data mapping and harmonization (★)<br>- D1.1: Map custom data formats to standard (e.g. FHIR) ([14])<br>- D1.2: Align differing data formats between the FL platform and third-party systems (★) |
| Step 2: Data curation (★)<br>- D2: Balance the ability of FL researchers to retrieve relevant on-site data with data privacy ([15, 22]) |
| Step 3: Data filtering and ensuring of DL-readiness (★)<br>- H3.1: Address unharmonized, low-quality data and annotations (★,[13, 15, 19, 20, 23, 30])<br>*→S3.1: Curate data carefully on FL platform by visual or metadata inspection (★)*<br>- H3.2: Resolve inconsistencies in data-annotation references (★,[13])<br>*→S3.2: Employ data validation workflows to verify data-annotation referencing, e.g. via StudyInstanceUID in DICOM (★)*<br>- H3.3: Handle data being corrupted or violating standards or specifications (★)<br>*→S3.3: Employ data validation workflows to verify all data attributes are suitable for processing by the DL algorithm, e.g. dimensionality, orientation, number of annotation labels (★)* |
| Step 4: Handling of data-on-algorithm issues (★)<br>- H4.1: DL algorithm fails on site's data (★)<br>*→S4.1: Debug DL algorithm on site's data with FL researchers and site IT (★)*<br>- H4.2: Malicious data identified solely through worse model performances at specific sites (★)<br>*→S4.2: Conduct manual inspection of data to identify invalid samples missed by data validation workflows (★,[10])* |
| **Phase 5: Experiments and evaluation** |
| Step 1: Running FL experiments (★)<br>- D1.1: Minimize strain on local hospital resources ([14])<br>- D1.2: Manage limited communication capacities ([8])<br>- D1.3: Manage resource availability if FL platform is used in multiple projects (★)<br>- D1.4: Delayed convergence of FL model due to data heterogeneity ([21]) |
| Step 2: Handling stragglers (★)<br>- D2.1: Varying durations of federated communication rounds due to heterogeneous hardware, network connections (★,[9, 21, 25, 27, 28]) and communication latencies ([21])<br>- D2.2: Idle machine times at faster sites due to stragglers (★) |
| Step 3: Managing failing FL experiments (★)<br>- D3.1: Failing FL experiments ([24, 27, 29]) caused by drop-outs of offline sites (★,[5, 21]), IT issues, and configurations like nightly VM backups, GPU driver mismatch (★)<br>- H3.1: Handle failing (([24, 27, 29]) or offline sites ([21]) resiliently (★)<br>*→S3.1: Save model checkpoints for restarts of failed FL experiments (★)*<br>- D3.2: Increased on-site debugging efforts due to error-prone experiments (★), insufficient error logging (★,[10]), limited technical documentation of FL platform ([10]) |
| Step 4: Restarting failed FL experiments (★)<br>- H4.1: Ensure site's readiness before restarting a FL experiment, including checking of logs, bugs, resources, and model checkpoints (★)<br>*→S4.1: Automatically share error logs*<br>*→S4.2: Communicate between FL researchers and site IT to resolve bugs, and verify availability of model checkpoints and resources (★)* |
| Step 5: Evaluate model performance on sites (★)<br>- H5: Have final FL model available (★)<br>*→S5.1: FL platform saves final FL model in one additional federated communication round (★)*<br>*→S5.2: Compress binaries of final FL model and send via communication channel, e.g. zip-file via mail (★)* |
| Step 6: Handling issues of test data on evaluation algorithm (★)<br>- D6.1: Test data filtering and ensuring of DL-readiness via data validation workflows (★)<br>- H6.2: Evaluation algorithm fails on site's test data (★)<br>*→S6.2: Debug evaluation algorithm on site's data to resolve issue with FL researchers and site IT (★)* |

## Real-world FL study: Experimental results

The experimental results demonstrate the functionality of our FL infrastructure within the RACOON project, developed by successfully overcoming the identified practical hurdles and difficulties. We investigated benefits hospitals gain by leveraging the power of FL, demonstrated across three distinct evaluation scenarios.

### Data characteristics

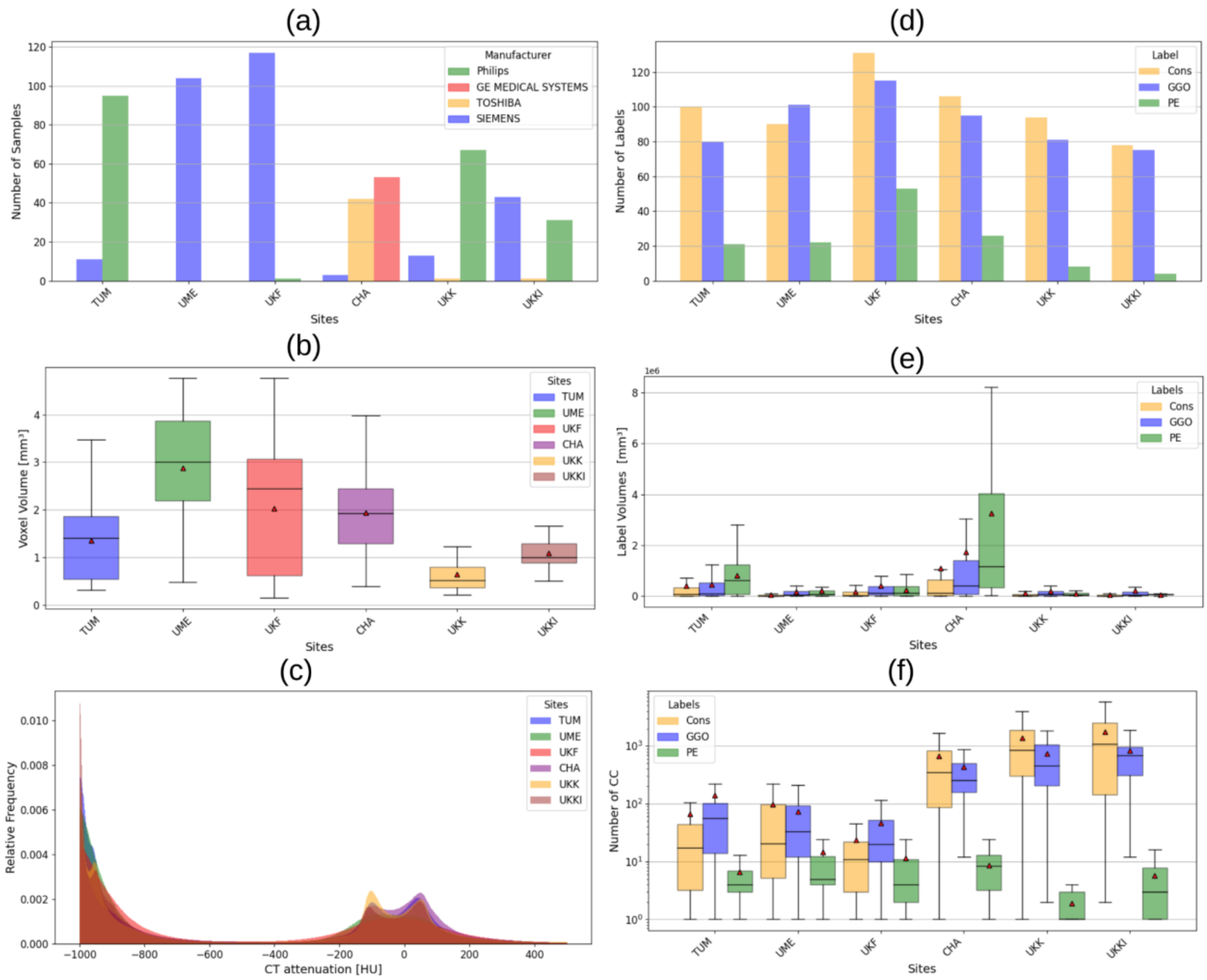

Figure 3: Data characteristics of CT data and annotations labels across the six participating sites. (a) CT scanner manufacturer distribution; (b) Average voxel volume distribution of CT scans; (c) Relative histogram of CT attenuation in HU (Note: -1000 HU visualizing air); (d) Annotation label distribution; (e) Annotation label volume distribution; (f) CCA: Number of CC per annotation label.

Data heterogeneity across participating sites impacts performance of models trained via FL[43]. We examined the characteristics of the distributed data to understand the degree of heterogeneity, see descriptive statistics in figure 3. Since CTs were provided pseudonymized, we relied on technical metadata and image-derived characteristics, excluding demographic details of the cohort.

The dataset comprises CT scans from four manufacturers, predominantly Siemens and Philips, with site CHA being an outlier. The voxel volumes vary from 0.15 (site UKK) to 4.84 $mm^3$ (site UME). The HU intensity distributions across the scans are similar due to normalization of HU values. Annotation labels, including PE as the least occurring pathology, are uniformly distributed across sites. We analyzed the distinctions between site's annotations by examining annotation volumes and conducting a connected component analysis (CCA). Site CHA features the largest annotation volumes, especially for PE cases. The CCA highlighted significant variations in annotation procedures among sites, with automatically pre-processed sites (CHA, UKK, UKKI) having a higher count of connected components (CC) compared to manually annotated sites (TUM, UME, UKF).

## Segmentation performance evaluation

### *Personalization*

Evaluating the personalization capabilities among manually annotated sites (TUM, UME, UKF) shows an overall superior performance of $Spec(FL_{man})$ achieving the best rank with average metrics: DSC = 0.47 (95% CI: 0.43-0.52), NSD = 0.39 (95% CI: 0.35-0.44), HSD = 127.94 mm (95% CI: 114.40-141.57) and NAVE = 6.20 (95% CI: 0.0-13.57). The introduced specialization stabilizes the segmentation performances across all metrics, whereas non-specialized models ($L_i$, $E_{man}$, $FL_{man}$) show considerable variability. We conclude that collaborative approaches outperform local models $L_i$, while among the non-specialized collaborative approaches $FL_{man}$ outperforms $E_{man}$ (figure 4.a, table A.2; qualitative results in figure 5).

Considering the personalization capabilities across automatically pre-processed sites (CHA, UKK, UKKI), we obtain superior performance of the specialized and non-specialized FL models ($Spec(FL_{auto})$ and $FL_{auto}$) on average rank with average metrics for $Spec(FL_{auto})$ of DSC = 0.41 (95% CI: 0.40-0.51), NSD = 0.40 (95% CI: 0.39-0.49), HSD = 127.96 mm (95% CI: 106.51-134.51), NAVE = 2.02 (95% CI: 0.49-3.23). Despite $Spec(FL_{auto})$ achieving only a single first-place ranking, its specialization contributes to a more consistent performance across metrics, securing its top position on average rank. Moreover, the results reveal a trend that sites with poor local model performance gain greater benefits from FL (figure 4.b, table A.3).

Incorporating all six sites in the benchmarking introduces a high data heterogeneity due to the differences in annotations. Despite this, we obtain for $Spec(FL_{all})$ the best average ranking with average metrics of DSC = 0.44 (95% CI: 0.43-0.50), NSD = 0.38 (95% CI: 0.36-0.43), HSD = 136.48 mm (95% CI: 122.60-143.33) and NAVE = 10.10 (95% CI: 0.0-35.81). The results support previously observed trends that specialization leads to more stable performances and ranks, while sites with poorer local performance notably benefit from FL (figure 4.c, table A.4).

We investigate the trade-off between enlarging the federation to six sites and thereby increasing data heterogeneity. By evaluating $Spec(FL_{all})$ on both manually and automatically pre-processed sites individually (table A.5), and comparing these results with $Spec(FL_{man})$ and $Spec(FL_{auto})$ (table A.2-A.3), we found that neither group of sites are benefitting in all metrics from the federation enlargement.

### *Generalization without local training*

The generalization performance is evaluated among the three manually annotated sites TUM, UME and UKF. The best generalizing model on average rank is $FL_{leave-i-out}$ with average metrics of DSC = 0.42 (95% CI: 0.37-0.46), NSD = 0.33 (95% CI: 0.29-0.37), HSD = 140.93 mm (95% CI: 127.30-155.77) and NAVE = 9.52 (95% CI: 0.0-23.97). The results reveal that local models $L_{i \neq j}$ from other sites $j$ do not generalize well, whereas collaborative approaches, particularly with $FL_{leave-i-out}$, demonstrate superior generalization capabilities (figure 4.d, table A.6).

### *Generalization with local training*

For sites with training capabilities seeking to circumvent efforts of real-world FL, the results consistently reveal that collaborative approaches are superior compared to local models $L_i$. Although non-specialized $E$ outperforms $FL_{leave-i-out}$, the top-performing model is $Spec(FL_{leave-i-out})$ with average metrics of DSC = 0.46 (95% CI: 0.41-0.50), NSD = 0.37 (95% CI: 0.33-0.41), HSD = 125.48 mm (95% CI: 112.06-139.92), NAVE = 3.65 (95% CI: 1.20-6.08). This suggests that the greatest benefit for site $i$ is achieved by adopting a FL model trained by other sites $j$ and specializing it with their local model $L_i$ (figure 4.e, table A.7).

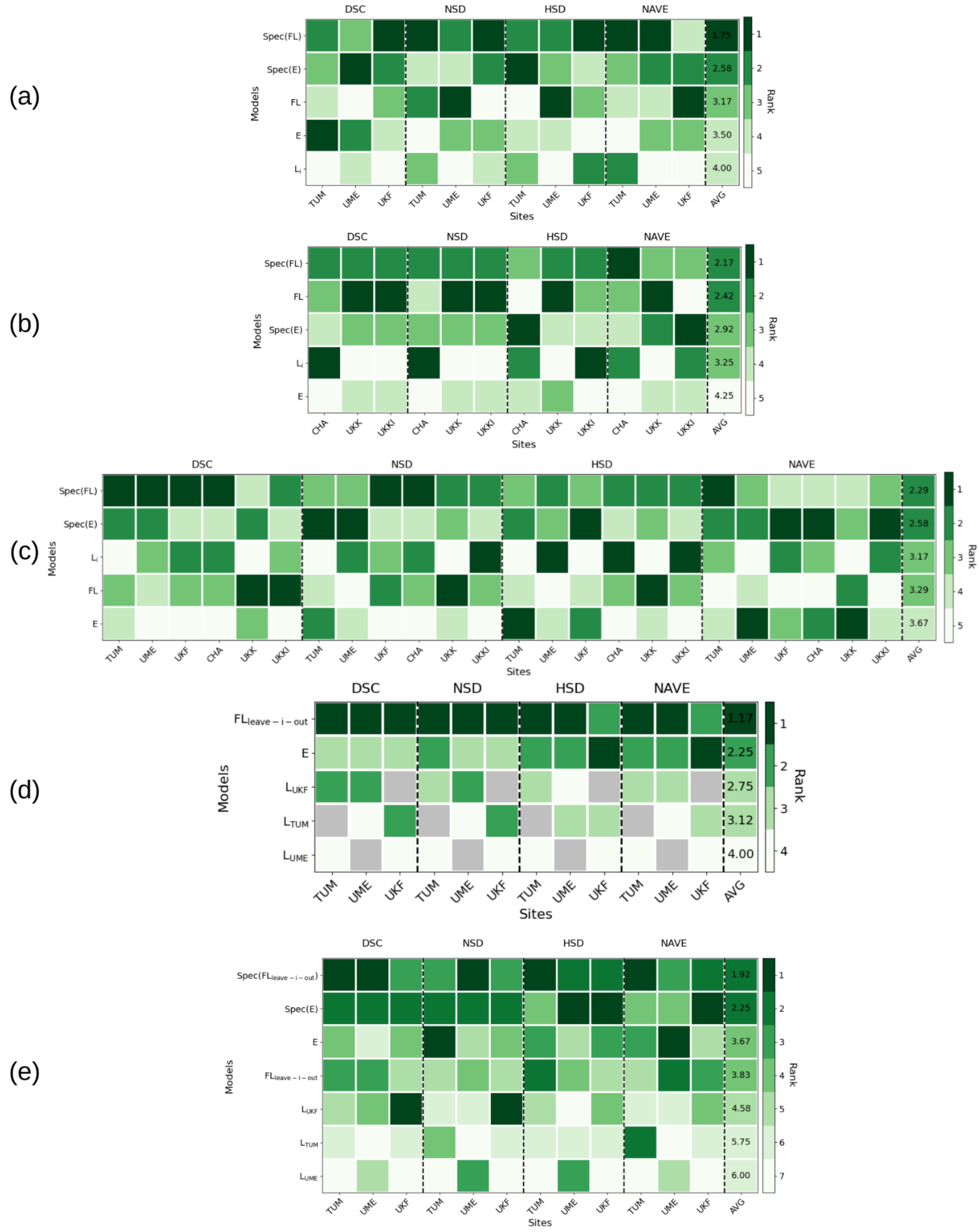

Figure 4: Heatmap visualization of the ranks achieved by compared models over the four metrics at participating sites including the average rank. (a) Personalization of manually annotated sites; (b) Personalization of automatically pre-processed sites; (c) Personalization of all sites; (d) Generalization without local training of manually annotated sites; (e) Generalization with local training of manually annotated sites.

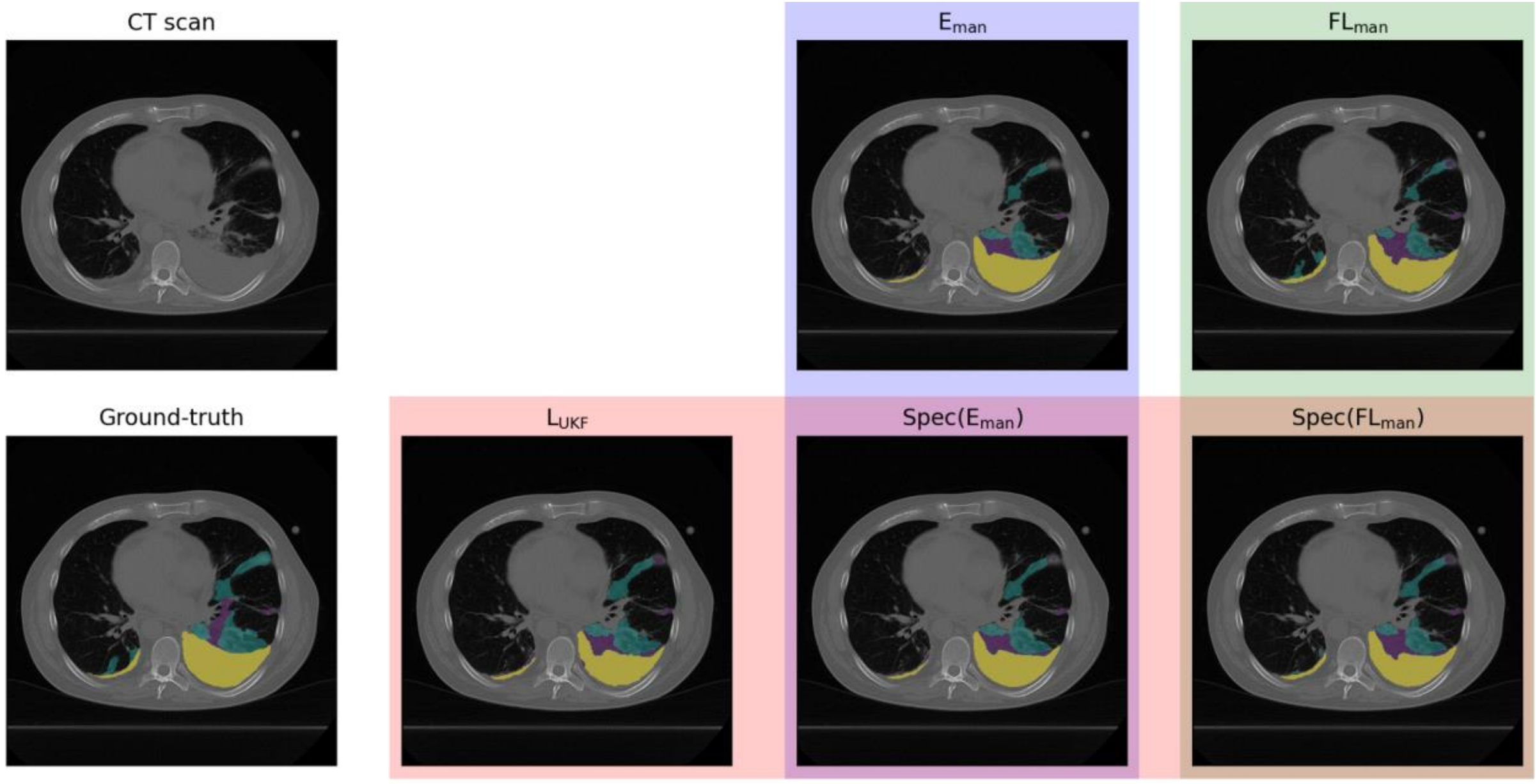

Figure 5: Qualitative segmentation results on a test sample from site UKF with Cons in violet, GGO in cyan, PE in yellow. Segmentation predictions of the models $L_{UKF}$, $E_{man}$, $FL_{man}$, with the specialization approach for $Spec(E_{man})$ and $Spec(FL_{man})$ highlighted.

# DISCUSSION

The guide we propose for building real-world FL initiatives details essential steps, identifies potential hurdles, and provides practical solutions that we have implemented, steering clear of hypothetical solutions. All issues associated with real-world FL were either resolved or circumvented, whereas key solutions included ensuring that each site had motivated and capable staff through targeted incentives. Additionally, defining data specifications during the organizational phase and employing data validators in the experiment preparation phase were crucial to ensure data-readiness for training and evaluation. Insufficient data specifications led to varied annotation procedures among radiologists, resulting in poorer model performance (figure 4.b). This was evident from the performance disparities observed between sites with manual versus automatically pre-processed annotations (table A.4- A.6). Our most effective solution involved providing offline-installable VMs to deploy FL platforms at sites with restricted internet access. Conversely, our least effective solution was lengthy troubleshooting sessions between FL platform-, third-party system engineers and on-site IT staff to debug software interfaces or deployed algorithms on site's data. Significantly hindering difficulties were idle machines at faster sites due to heterogeneous infrastructure and frequent errors during the initial experimental phase until we identified and addressed contributing factors like nightly backups and other maintenance activities conducted by local IT.

In our experimental studies, we first analyzed data characteristics to assess the heterogeneity of real-world data. This proved invaluable, as it enabled us to detect differences in annotation procedures using a CCA, highlighting a significant source of data heterogeneity. In terms of segmentation performance, collaborative approaches – $E$, $Spec(E)$, $FL$, and $Spec(FL)$ – consistently outperform local models $L_i$ in all evaluation scenarios, highlighting the power of collaborative training. Among these, the $FL$ model and the specialized $Spec(FL)$ showed superior performance over ensembling approaches in both personalization and generalization scenarios. These results demonstrate that despite the substantial effort involved in conducting real-world FL, the benefits clearly justify the investment. Moreover, our experimental studies demonstrate that randomly adding participating sites to a federation can worsen model performances due to increased data heterogeneity.

The proposed guide presents a starting point that will not cover all potential issues and is meant to be extended by future real-world FL efforts. Additionally, although we integrated a broad spectrum of FL literature, some insights might be specific to our initiative and may not universally apply. The experimental results are influenced by various factors: data heterogeneity from varying annotation procedures, choosing the default federated averaging as aggregation strategy, and using a single fold in nnU-Net rather than a five-fold cross-validation. Efficiency-driven decisions, such as using nnU-Net's low-resolution model and training for fixed 1000 epochs, also affected performance. Lastly, the decisions to select the challenging lung pathology segmentation task[39], and to include sites with low segmentation performances on own test data, may not have been ideal for evaluating FL in real-world conditions.

Despite its limitations, our study outlines possibilities for future research. With the established FL infrastructure in the RACOON initiative, we are now equipped to investigate clinically relevant research questions at scale, leveraging the power of FL. From a FL perspective, exploring client selection emerges as a promising area of research, particularly given the observed impact of heterogeneous annotations on model performance. This could lead to development of automatic proxies pre-evaluating a site’s participation in FL training. Moreover, respecting the lower effort and strong performance of ensemble approaches compared to local models, further explorations how ensemble approaches can enhance personalization and generalization are advantageous.

## CONCLUSION

In this work, we strive towards bridging the gap between simulated and real-world FL research. We identified a significant gap in literature of detailed insights into successfully establishing FL initiatives. To address this, we developed a FL initiative within the German RACOON project and conducted real-world experiments demonstrating that FL models consistently outperformed compared approaches across all evaluation scenarios. We compiled insights from both our own experiences and literature into a comprehensive guide. This guide details necessary steps, encountered issues and suggests solutions involved in building and deploying real-world FL. With this guide we aim to streamline and accelerate future real-world FL initiatives by guiding through the development process and to help avoiding pitfalls. Furthermore, we aim to contribute a further step towards our ultimate objective, the widespread use of FL in clinical applications to improve patient diagnosis and therapy through powerful and efficient models trained collaboratively on distributed data.

## Acknowledgements

The following individuals contributed significant work to the development of the Kaapana platform:

- Jonas Scherer (German Cancer Research Center Heidelberg)
- Klaus Kades (German Cancer Research Center Heidelberg)
- Ralf Floca (German Cancer Research Center Heidelberg)
- Marco Nolden (German Cancer Research Center Heidelberg)
- Hanno Gao (German Cancer Research Center Heidelberg)
- Philipp Schader (German Cancer Research Center Heidelberg)
- Santhosh Parampottupadam (German Cancer Research Center Heidelberg)
- Lorenz Feineis (German Cancer Research Center Heidelberg)
- Jens Beyermann (German Cancer Research Center Heidelberg)
- Benjamin Hamm (German Cancer Research Center Heidelberg)
- Rajesh Baidya (German Cancer Research Center Heidelberg)
- Mikulas Bankovic (German Cancer Research Center Heidelberg)

Additionally, the following individuals supported the project at the participating sites:

- Leonhard Feiner (TU Munich)
- Enrico Nasca (University Hospital Essen)
- Jonathan Kottlors (University Hospital Cologne)
- Benedikt Wichtlhuber (University Hospital Frankfurt)

## Funding

This work was funded by „NUM 2.0“(FKZ: 01KX2121).

## Supplementary Materials

Table A.1: nnU-Net parameters of models $FL_{man}$, $FL_{auto}$ and $FL_{all}$.

| **model** | **$FL_{man}$** | | **$FL_{auto}$** | | **$FL_{all}$** | |
|---|---|---|---|---|---|---|
| **parameters** | **stage 0** | **stage 1** | **stage 0** | **stage 1** | **stage 0** | **stage 1** |
| **Batch size** | 2 | 2 | 2 | 2 | 2 | 2 |
| **Number of pooling per axis** | [4, 5, 5] | [4, 5, 5] | [4, 5, 5] | [4, 5, 5] | [4, 5, 5] | [4, 5, 5] |
| **Patch size** | [112, 160, 128] | [112, 160, 128] | [96, 160, 160] | [96, 160, 160] | [112, 160, 128] | [112, 160, 128] |
| **Median image shape in voxels** | [165, 233, 233] | [361, 512, 512] | [157, 250, 250] | [322, 512, 512] | [164, 233, 233] | [164, 233, 233] |
| **Original spacing [mm]** | [0.90, 0.77, 0.77] | [0.90, 0.77, 0.77] | [1.00, 0.72, 0.72] | [1.00, 0.72, 0.72] | [0.90, 0.75, 0.75] | [0.90, 0.75, 0.75] |
| **Target spacing [mm]** | [1.98, 1.70, 1.70] | [0.90, 0.77, 0.77] | [2.05, 1.47, 1.47] | [1.00, 0.72, 0.72] | [1.98, 1.64, 1.64] | [0.90, 0.75, 0.75] |
| **Pooling operation kernel sizes** | [[2, 2, 2], [2, 2, 2], [2, 2, 2], [2, 2, 2], [1, 2, 2]] | [[2, 2, 2], [2, 2, 2], [2, 2, 2], [2, 2, 2], [1, 2, 2]] | [[2, 2, 2], [2, 2, 2], [2, 2, 2], [2, 2, 2], [1, 2, 2]] | [[2, 2, 2], [2, 2, 2], [2, 2, 2], [2, 2, 2], [1, 2, 2]] | [[2, 2, 2], [2, 2, 2], [2, 2, 2], [2, 2, 2], [1, 2, 2]] | [[2, 2, 2], [2, 2, 2], [2, 2, 2], [2, 2, 2], [1, 2, 2]] |
| **Convolution operation kernel sizes** | [[3, 3, 3], [3, 3, 3], [3, 3, 3], [3, 3, 3], [3, 3, 3], [3, 3, 3]] | [[3, 3, 3], [3, 3, 3], [3, 3, 3], [3, 3, 3], [3, 3, 3], [3, 3, 3]] | [[3, 3, 3], [3, 3, 3], [3, 3, 3], [3, 3, 3], [3, 3, 3], [3, 3, 3]] | [[3, 3, 3], [3, 3, 3], [3, 3, 3], [3, 3, 3], [3, 3, 3], [3, 3, 3]] | [[3, 3, 3], [3, 3, 3], [3, 3, 3], [3, 3, 3], [3, 3, 3], [3, 3, 3]] | [[3, 3, 3], [3, 3, 3], [3, 3, 3], [3, 3, 3], [3, 3, 3], [3, 3, 3]] |

Table A.2: Personalization segmentation evaluation performance among sites with manually generated annotations (TUM, UME, UKF).

| | **test data** | | | | | | | | | | | | **avg** | | | | **rank** |
|---|---|---|---|---|---|---|---|---|---|---|---|---|---|---|---|---|---|
| | **TUM** | | | | **UME** | | | | **UKF** | | | | | | | | |
| | **DSC** | **NSD** | **HSD** | **NAVE** | **DSC** | **NSD** | **HSD** | **NAVE** | **DSC** | **NSD** | **HSD** | **NAVE** | **DSC** | **NSD** | **HSD** | **NAVE** | |
| **$L_i$** | 0,44 | 0,33 | 156,53 | 2,04 | 0,41 | 0,36 | 117,99 | 8,56 | 0,50 | 0,40 | 146,48 | 43,69 | 0,45 | 0,36 | 140,33 | 18,09 | 4,00 |
| **E** | 0,47 | 0,34 | 147,47 | 2,71 | 0,43 | 0,37 | 119,23 | 8,53 | 0,43 | 0,36 | 131,35 | 3,65 | 0,44 | 0,36 | 132,69 | 4,96 | 3,50 |
| **FL** | 0,46 | 0,32 | 146,64 | 1,82 | 0,44 | 0,35 | 155,62 | 4,20 | 0,49 | 0,39 | 133,18 | 2,32 | 0,46 | 0,36 | 145,14 | **2,78** | 3,17 |
| **Spec(E)** | 0,47 | 0,34 | 144,50 | 2,60 | 0,42 | 0,38 | 104,75 | 8,43 | 0,46 | 0,39 | 123,96 | 3,06 | 0,45 | 0,37 | **124,41** | 4,70 | 2,58 |
| **Spec(FL)** | 0,47 | 0,33 | 144,33 | 1,36 | 0,43 | 0,39 | 117,85 | 6,56 | 0,51 | 0,45 | 121,65 | 10,66 | **0,47** | **0,39** | 127,94 | 6,20 | **1,75** |

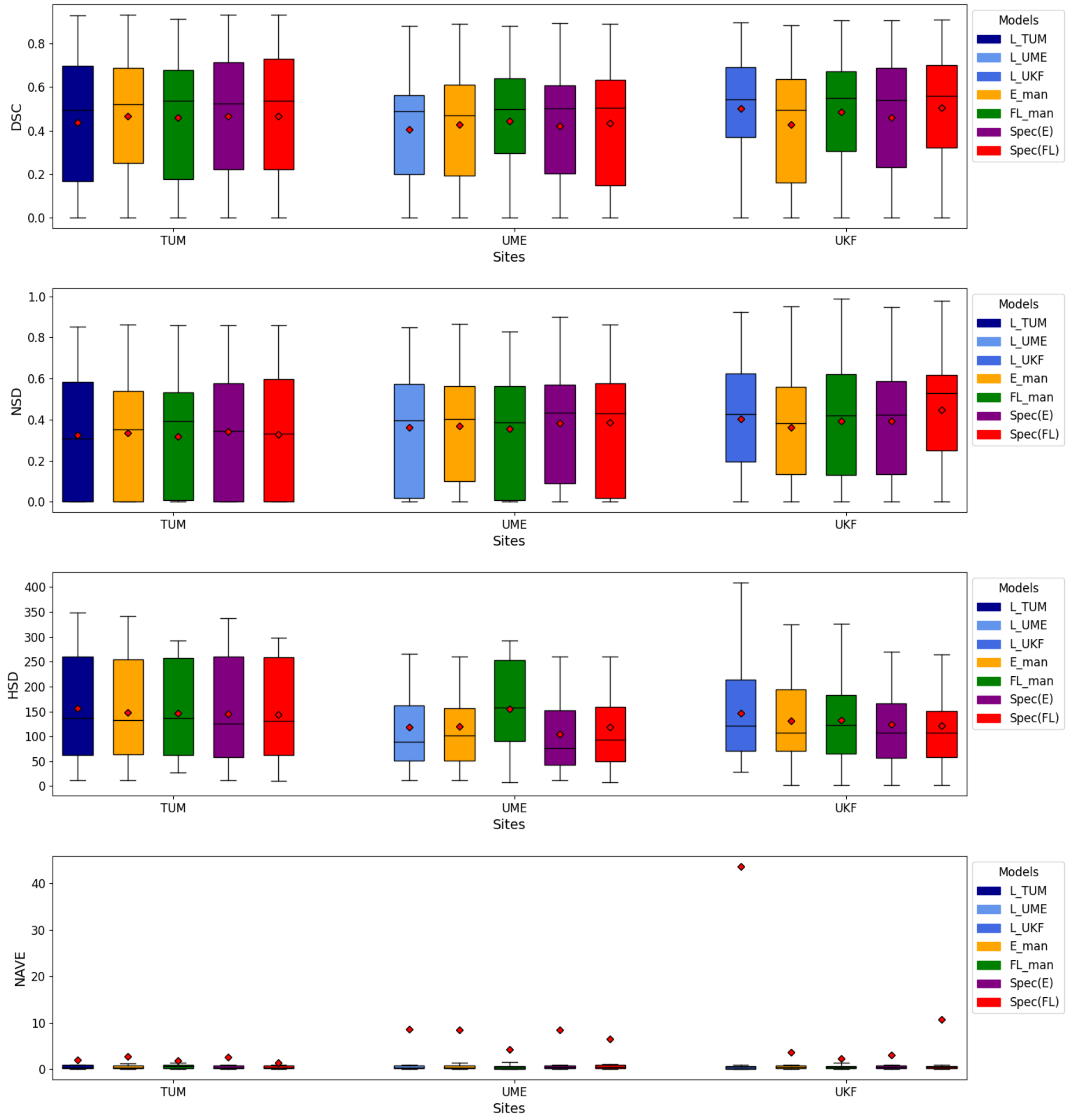

Figure A.1: Boxplots of personalization segmentation evaluation performance among sites with manually generated annotations (TUM, UME, UKF) of metrics DSC, NSD, HSD and NAVE.

Table A.3: Personalization segmentation evaluation performance among sites with automatically pre-processed annotations (CHA, UKK, UKKI).

| | **test data** | | | | | | | | | | | | avg | | | | **rank** |
|---|---|---|---|---|---|---|---|---|---|---|---|---|---|---|---|---|---|
| | **CHA** | | | | **UKK** | | | | **UKKI** | | | | | | | | |
| | **DSC** | **NSD** | **HSD** | **NAVE** | **DSC** | **NSD** | **HSD** | **NAVE** | **DSC** | **NSD** | **HSD** | **NAVE** | **DSC** | NSD | HSD | NAVE | |
| **$L_i$** | 0,60 | 0,57 | 100,20 | 1,54 | 0,08 | 0,07 | 214,80 | 4,91 | 0,36 | 0,37 | 115,25 | 1,84 | 0,34 | 0,34 | 143,42 | 2,76 | 3,08 |
| **E** | 0,46 | 0,49 | 107,17 | 2,65 | 0,24 | 0,20 | 176,90 | 2,70 | 0,36 | 0,36 | 135,85 | 2,15 | 0,35 | 0,35 | 139,97 | 2,50 | 4,25 |
| **FL** | 0,55 | 0,52 | 111,97 | 1,62 | 0,39 | 0,34 | 139,86 | 0,49 | 0,38 | 0,35 | 125,92 | 3,66 | **0,44** | 0,40 | 125,92 | 1,92 | 2,75 |
| **Spec(E)** | 0,53 | 0,53 | 95,39 | 1,68 | 0,26 | 0,24 | 182,74 | 2,29 | 0,36 | 0,37 | 127,70 | 1,79 | 0,39 | 0,38 | 135,28 | 1,92 | 2,83 |
| **Spec(FL)** | 0,59 | 0,56 | 101,44 | 1,45 | 0,27 | 0,27 | 163,04 | 2,68 | 0,37 | 0,37 | 119,41 | 1,92 | 0,41 | 0,40 | 127,96 | 2,02 | **2,08** |

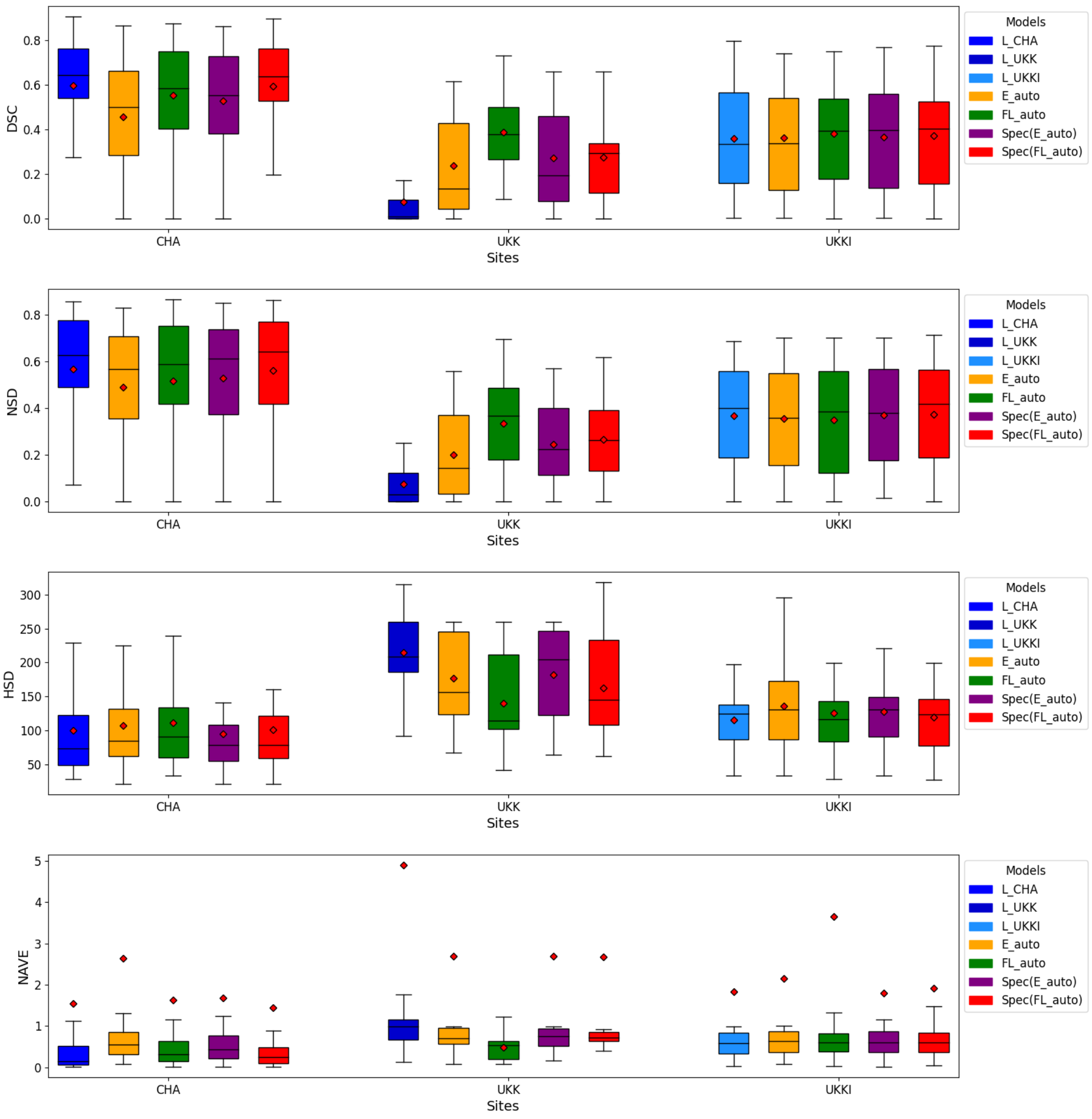

Figure A.2: Boxplots of personalization segmentation evaluation performance among sites with automatically pre-processed annotations (CHA, UKK, UKKI) of metrics DSC, NSD, HSD and NAVE.

Table A.4: Personalization segmentation evaluation performance among all sites.

| | test data | | | | | | | | | | | | | | | | | | | | | | | | avg | | | | rank |
|---|---|---|---|---|---|---|---|---|---|---|---|---|---|---|---|---|---|---|---|---|---|---|---|---|---|---|---|---|---|
| | TUM | | | | UME | | | | UKF | | | | CHA | | | | UKK | | | | UKKI | | | | | | | | |
| | DSC | NSD | HSD | NAVE | DSC | NSD | HSD | NAVE | DSC | NSD | HSD | NAVE | DSC | NSD | HSD | NAVE | DSC | NSD | HSD | NAVE | DSC | NSD | HSD | NAVE | DSC | NSD | HSD | NAVE | |
| **$L_i$** | 0,44 | 0,33 | 156,53 | 2,04 | 0,41 | 0,36 | 117,99 | 8,56 | 0,50 | 0,40 | 146,48 | 43,69 | 0,60 | 0,57 | 100,20 | 1,54 | 0,08 | 0,07 | 214,80 | 4,91 | 0,36 | 0,37 | 115,25 | 1,84 | 0,40 | 0,35 | 141,88 | 10,43 | 3,17 |
| **E** | 0,45 | 0,36 | 144,22 | 2,04 | 0,39 | 0,35 | 143,63 | 5,00 | 0,40 | 0,35 | 127,44 | 44,74 | 0,54 | 0,48 | 122,92 | 1,51 | 0,29 | 0,19 | 192,66 | 0,59 | 0,33 | 0,25 | 183,49 | 3,18 | 0,40 | 0,33 | 152,39 | 9,51 | 3,67 |
| **FL** | 0,46 | 0,33 | 154,07 | 2,97 | 0,40 | 0,32 | 164,59 | 7,35 | 0,47 | 0,41 | 145,15 | 138,48 | 0,60 | 0,53 | 108,78 | 3,77 | 0,39 | 0,28 | 149,02 | 0,61 | 0,40 | 0,31 | 145,31 | 16,29 | **0,45** | 0,36 | 144,49 | 28,24 | 3,29 |
| **Spec(E)** | 0,46 | 0,36 | 145,24 | 1,79 | 0,41 | 0,36 | 128,10 | 5,64 | 0,43 | 0,37 | 126,60 | 6,73 | 0,56 | 0,51 | 109,91 | 1,37 | 0,30 | 0,19 | 182,49 | 0,63 | 0,34 | 0,28 | 162,96 | 1,29 | 0,42 | 0,35 | 142,55 | **2,91** | 2,58 |
| **Spec(FL)** | 0,46 | 0,34 | 148,15 | 1,37 | 0,42 | 0,36 | 126,06 | 6,18 | 0,51 | 0,45 | 129,80 | 48,29 | 0,61 | 0,58 | 101,55 | 1,86 | 0,28 | 0,22 | 168,17 | 0,68 | 0,37 | 0,31 | 145,14 | 2,24 | 0,44 | **0,38** | **136,48** | 10,10 | **2,29** |

Table A.5: Personalization segmentation evaluation performances among all sites averaged over all sites (all), over manual (man) and automatic (auto) sites.

| | avg man | | | | avg auto | | | |
|---|---|---|---|---|---|---|---|---|
| | DSC | NSD | HSD | NAVE | DSC | NSD | HSD | NAVE |
| **$L_i$** | 0,45 | 0,36 | 140,33 | 18,09 | 0,34 | 0,34 | 143,42 | 2,76 |
| **E** | 0,41 | 0,35 | 138,43 | 17,26 | 0,39 | 0,30 | 166,36 | 1,76 |
| **FL** | 0,44 | 0,35 | 154,61 | 49,60 | **0,46** | **0,37** | **134,37** | 6,89 |
| **Spec(E)** | 0,43 | 0,36 | **133,31** | **4,72** | 0,40 | 0,33 | 151,78 | **1,10** |
| **Spec(FL)** | **0,47** | **0,38** | 134,67 | 18,61 | 0,42 | **0,37** | 138,29 | 1,59 |

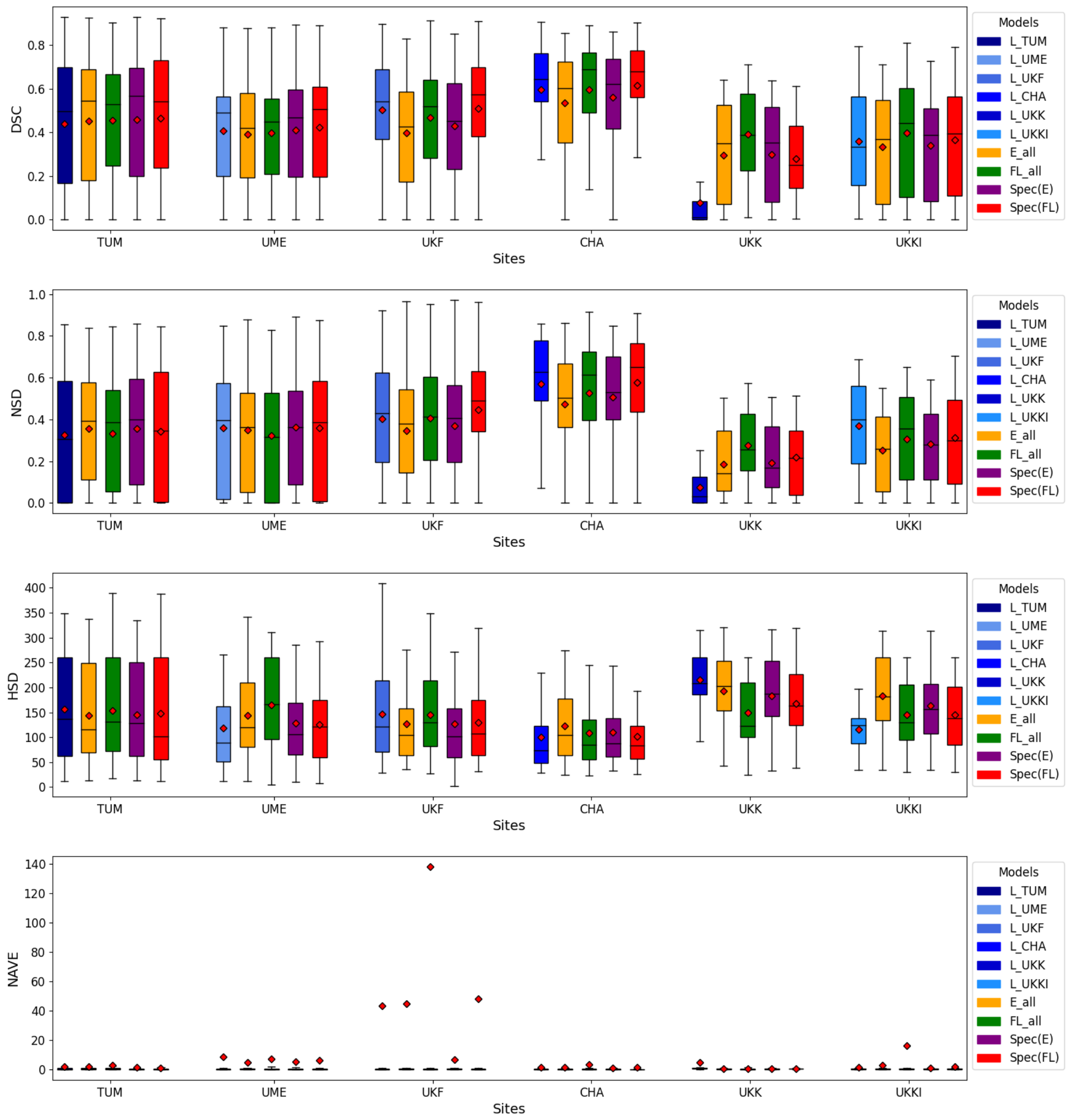

Figure A.3: Boxplots of personalization segmentation evaluation performance among all sites of metrics DSC, NSD, HSD and NAVE.

Table A.6: Generalization without local training segmentation evaluation performance among sites with manually generated annotations (TUM, UME, UKF).

| | test data | | | | | | | | | | | | avg | | | | rank |
|---|---|---|---|---|---|---|---|---|---|---|---|---|---|---|---|---|---|
| | TUM | | | | UME | | | | UKF | | | | | | | | |
| | DSC | NSD | HSD | NAVE | DSC | NSD | HSD | NAVE | DSC | NSD | HSD | NAVE | DSC | NSD | HSD | NAVE | |
| $L_{TUM}$ | | | | | 0,32 | 0,24 | 149,38 | 11,47 | 0,35 | 0,27 | 149,79 | 73,06 | 0,33 | 0,25 | 149,59 | 42,27 | 3,25 |
| $L_{UME}$ | 0,38 | 0,29 | 159,35 | 5,47 | | | | | 0,30 | 0,25 | 169,78 | 875,40 | 0,34 | 0,27 | 164,57 | 440,44 | 4,00 |
| $L_{UKF}$ | 0,44 | 0,29 | 151,58 | 4,11 | 0,42 | 0,33 | 155,31 | 9,20 | | | | | **0,43** | 0,31 | 153,45 | **6,65** | 2,75 |
| $E_{leave-i-out}$ | 0,44 | 0,30 | 142,26 | 3,56 | 0,39 | 0,32 | 137,82 | 8,17 | 0,34 | 0,29 | 143,21 | 12,06 | 0,39 | 0,30 | 141,10 | 7,93 | 2,17 |
| $FL_{leave-i-out}$ | 0,45 | 0,31 | 138,78 | 3,47 | 0,42 | 0,36 | 134,87 | 6,40 | 0,38 | 0,32 | 149,14 | 18,71 | 0,42 | **0,33** | **140,93** | 9,52 | **1,17** |

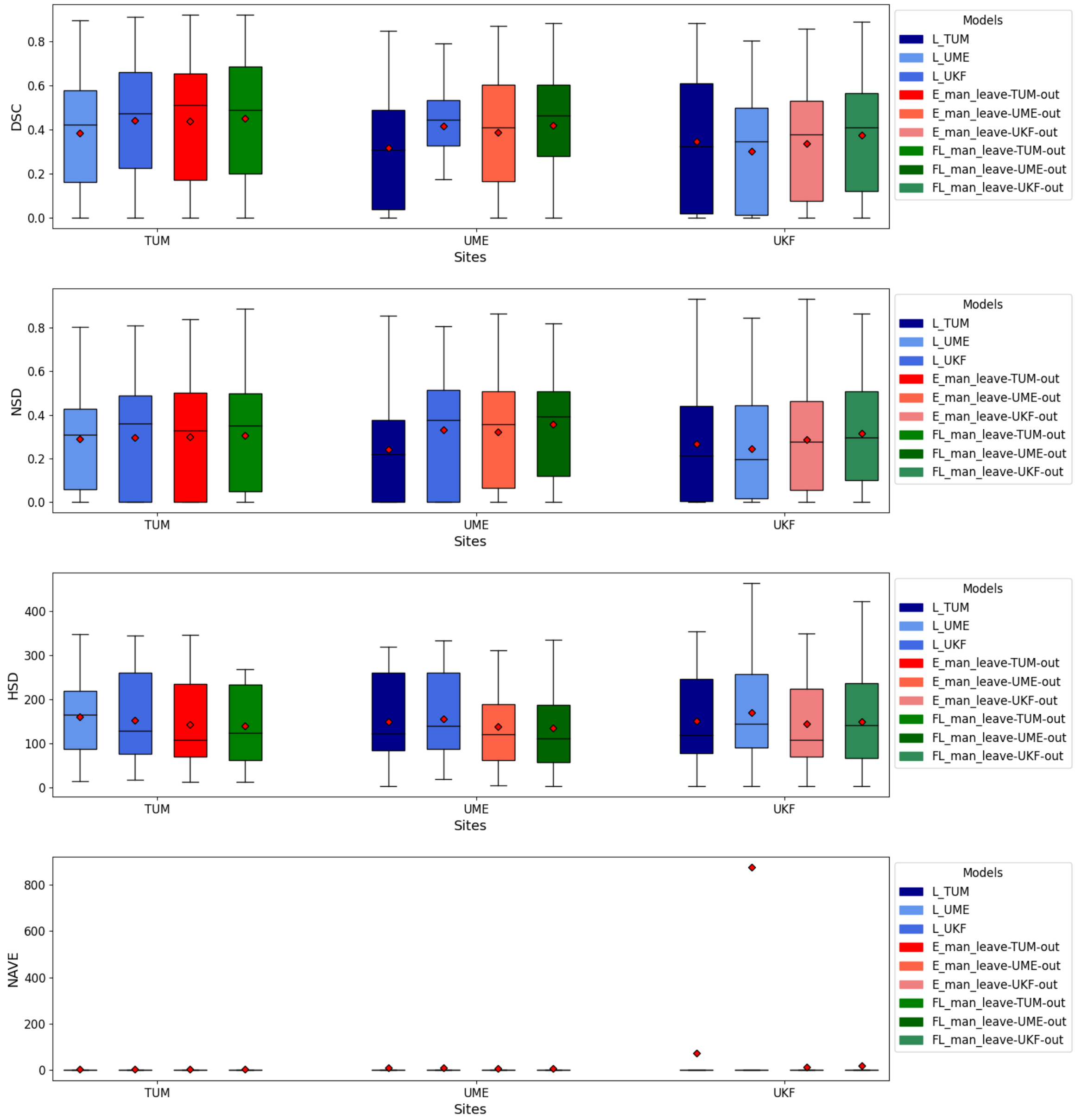

Figure A.4: Boxplots of generalization without local training segmentation evaluation performance among sites with manually generated annotations (TUM, UME, UKF) of metrics DSC, NSD, HSD and NAVE..

Table A.7: Generalization with local training segmentation evaluation performance among sites with manually generated annotations (TUM, UME, UKF).

| | **test data** | | | | | | | | | | | | **avg** | | | | **rank** |
|---|---|---|---|---|---|---|---|---|---|---|---|---|---|---|---|---|---|
| | **TUM** | | | | **UME** | | | | **UKF** | | | | | | | | |
| | **DSC** | **NSD** | **HSD** | **NAVE** | **DSC** | **NSD** | **HSD** | **NAVE** | **DSC** | **NSD** | **HSD** | **NAVE** | **DSC** | **NSD** | **HSD** | **NAVE** | |
| **$L_{TUM}$** | 0,44 | 0,33 | 156,53 | 2,04 | 0,32 | 0,24 | 149,38 | 11,47 | 0,35 | 0,27 | 149,79 | 73,06 | 0,37 | 0,28 | 151,90 | 28,86 | 5,73 |
| **$L_{UME}$** | 0,38 | 0,29 | 159,35 | 5,47 | 0,41 | 0,36 | 117,99 | 8,56 | 0,30 | 0,25 | 169,78 | 875,40 | 0,36 | 0,30 | 149,04 | 296,48 | 5,53 |
| **$L_{UKF}$** | 0,44 | 0,29 | 151,58 | 4,11 | 0,42 | 0,33 | 155,31 | 9,20 | 0,50 | 0,40 | 146,48 | 43,69 | 0,45 | 0,34 | 151,12 | 19,00 | 4,93 |
| **E** | 0,45 | 0,36 | 144,22 | 2,04 | 0,39 | 0,35 | 143,63 | 5,00 | 0,40 | 0,35 | 127,44 | 44,74 | 0,41 | 0,35 | 138,43 | 17,26 | 3,33 |
| **$FL_{leave-i-out}$** | 0,45 | 0,31 | 138,78 | 3,47 | 0,42 | 0,36 | 134,87 | 6,40 | 0,38 | 0,32 | 149,14 | 18,71 | 0,42 | 0,33 | 140,93 | 9,52 | 3,87 |
| **Spec(E)** | 0,47 | 0,34 | 144,50 | 2,60 | 0,42 | 0,38 | 104,75 | 8,43 | 0,46 | 0,39 | 123,96 | 3,06 | 0,45 | **0,37** | 124,41 | 4,70 | 2,33 |
| **Spec($FL_{leave-i-out}$)** | 0,48 | 0,33 | 136,57 | 0,98 | 0,44 | 0,39 | 112,69 | 6,79 | 0,45 | 0,39 | 127,19 | 3,19 | **0,46** | **0,37** | **125,48** | **3,65** | **2,27** |

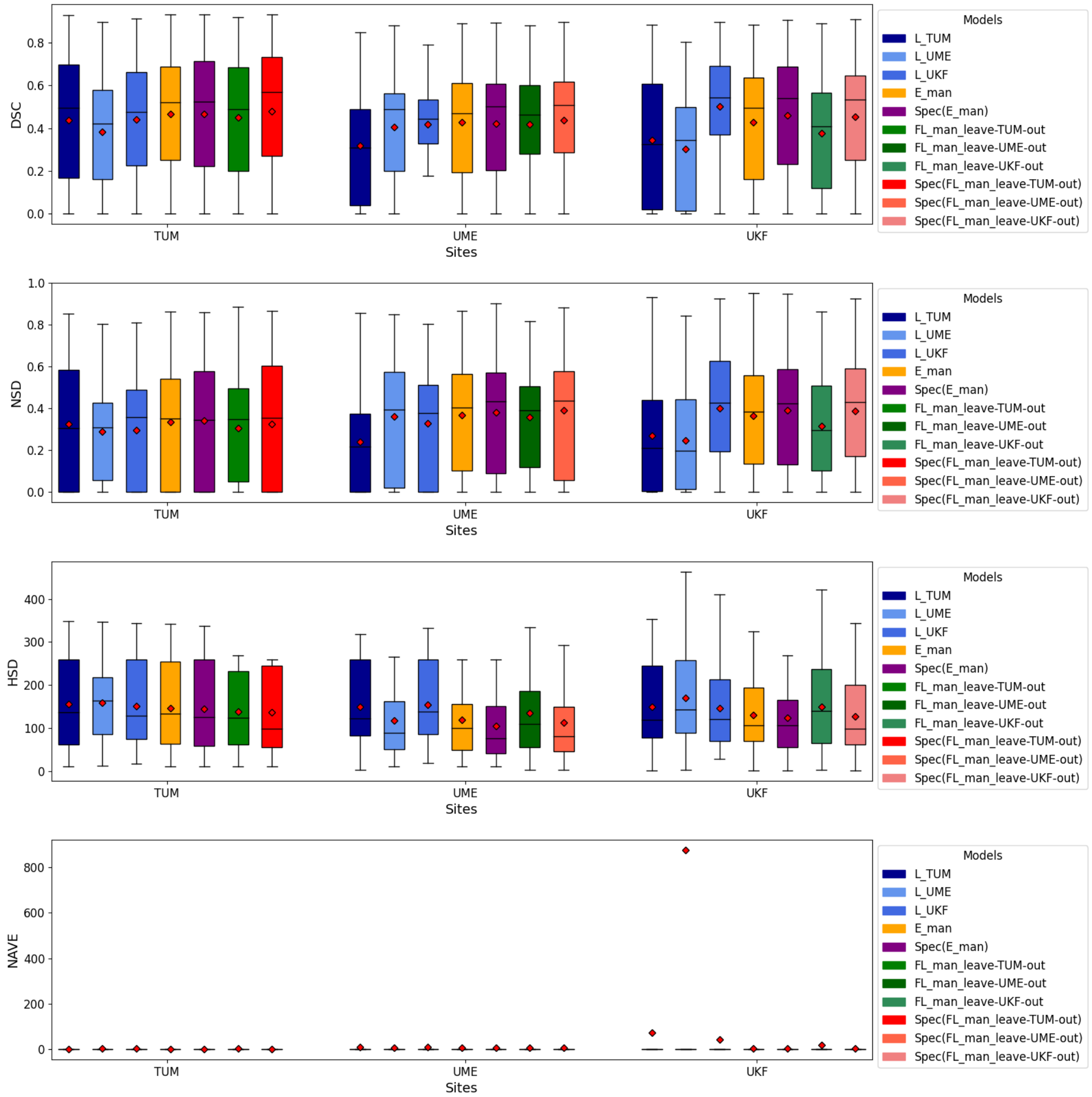

Figure A.5: Boxplots of generalization without local training segmentation evaluation performance among sites with manually generated annotations (TUM, UME, UKF) of metrics DSC, NSD, HSD and NAVE.